\documentclass[pdflatex,sn-mathphys-num, iicol]{sn-jnl}

\usepackage{graphicx}%
\usepackage{multirow}%
\usepackage{amsmath,amssymb,amsfonts}%
\usepackage{amsthm}%
\usepackage{mathrsfs}%
\usepackage[title]{appendix}%
\usepackage{xcolor}%
\usepackage{textcomp}%
\usepackage{manyfoot}%
\usepackage{booktabs}%
\usepackage{algorithm}%
\usepackage{algorithmicx}%
\usepackage{algpseudocode}%
\usepackage{listings}%
\usepackage{tabularx}
\usepackage{booktabs} 
\usepackage{subcaption}
\usepackage{float}
\usepackage{xr}

\theoremstyle{thmstyleone}%
\theoremstyle{thmstyletwo}%

\theoremstyle{thmstylethree}%

\begin{document}

\title[Article Title]{Multimodal Attention-based Deep Learning for Emergency Triage with Electronic Health Records}


\author[1]{\fnm{Hazqeel Afyq Athaillah} \sur{Kamarul Aryffin}}\email{hazqeel@student.usm.my}

\author*[2]{\fnm{Kamarul Aryffin} \sur{Baharuddin}}\email{amararyff@usm.my}

\author*[1,3]{\fnm{Mohd Halim} \sur{Mohd Noor}}\email{halimnoor@usm.my}

\affil[1]{\orgdiv{School of Computer Sciences}, \orgname{Universiti Sains Malaysia}, \orgaddress{\street{Jalan Universiti}, \city{Gelugor}, \postcode{11800}, \state{Pulau Pinang}, \country{Malaysia}}}

\affil[2]{\orgdiv{School of Medical Sciences}, \orgname{Universiti Sains Malaysia}, \orgaddress{\street{Jalan Raja Perempuan Zainab 2}, \city{Kubang Kerian}, \postcode{16150}, \state{Kelantan}, \country{Malaysia}}}

\affil[3]{\orgdiv{Artificial Intelligence Research Center}, \orgname{Ajman University}, \orgaddress{\street{University Street}, \city{Al Jurf 1}, \postcode{ 346}, \state{Ajman}, \country{United Arab Emirates}}}


\abstract{Accurate emergency triage decision is critical to avoid clinical deterioration, morbidity, and mortality. Machine learning-based triage system involves acquiring the main presenting complaint in text form and assessing vital signs in numerical data, enabling an automated and efficient analysis of patient information for timely and accurate prioritization of medical attention. However, modelling the intricacies of both data types requires a comprehensive understanding of the temporal structure and dependencies within the data. Thus, the aim of this study is to propose a multimodal deep learning architecture that can effectively handle both tabular and textual data. Furthermore, the proposed model exploits self-attention to to capture both local and global relationships between the features. A dataset consisting of 11,102 triage data collected from emergency department of Hospital Universiti Sains Malaysia is used for model development and validation. The proposed model demonstrated an increase of 1.95\% in accuracy, 2.49\% in F1-score, and 1.41\% in ROC AUC compared to the baseline model. The experimental results demonstrated the potential of the proposed model in predicting triage decisions.}

\keywords{Attention, Text Data, Tabular Data}



\maketitle

\section{Introduction}\label{sec1}

Triage is patients' first point of contact to the emergency department (ED) and hospital \cite{fry_review_2002}. Those who are hemodynamically unstable will be treated first, regardless of the time of arrival. This differs from the clinic settings where the visit is planned with an appointment, and patients are called based on the arrival time. The primary objective of triage is to rapidly identify patients with critical and time-sensitive conditions and to prioritize their care above those who can wait. At triage, patients are categorized to zones by trained nurses or medical assistants based on their vital signs and chief complaints. In some cases, patients were under-triaged, which may result in morbidity and mortality \cite{joseph_use_2023, tam_review_2018}. However, triage manual assessment is inherently subjective and highly prone to human error, as it relies on the individual experience, expertise and interpretation of the healthcare professionals. Therefore, there is a need for artificial intelligence (AI) based tools to help them ensure consistent and accurate decisions during the triage process.

Machine learning is an AI approach that have been used to predict various factors in medicine, including risk stratification, diagnosis, and treatment choice \cite{kumar_artificial_2023, javaid_significance_2022, davenport_potential_2019}. When built with algorithms and clinical workflow in mind, machine learning in the ED for augmented decision will improve care, decrease errors, increase efficiency, and eventually reduce morbidity and mortality \cite{raita_emergency_2019}. So far, the intersection of machine learning and emergency medicine has gathered much attention despite both areas having sophisticated development and rich individual sub-fields \cite{pavaloaia_artificial_2023, mueller_artificial_2022}.

During the triage process, the patient's vital signs, such as blood pressure, heart rate, and temperature, are numerical data, but the chief complaint from the patient is textual data. The patient's chief complaint is the most challenging as there is no standardized format for chief complaints. Furthermore, in machine learning, text data needs to be pre-processed because the data cannot be directly modeled using traditional machine learning algorithms such as support vector machine and decision tree \cite{maslej-kresnakova_comparison_2020}. Consequently, multimodal deep learning approaches have been proposed to effectively integrate structured tabular data with unstructured textual data into a unified predictive model. 

In \cite{leung_novel_2021, lin2024interpretable}, deep learning models that process both tabular and textual data is proposed for predicting patient hospitalization. The model makes use of Bidirectional Encoder Representations from Transformers (BERT) to produce text embedding and deep tabular network \cite{arik_tabnet_2020} to produce embedding for vital signs. Then, both embeddings are transformed and concatenated to produce image-like vector. The model utilizes convolutional layers to extract features from the image-like vector to produce prediction. Although convolution operations are able to extract salient local features, it is limited to understanding global context. Furthermore, reshaping the concatenated embeddings into grid matrices to accommodate standard 2D convolution operation introduces a rigid spatial dependencies that misalign with the heterogeneous triage data.

Therefore, this study aims to overcome the limitations of localized convolutional networks by developing a global, self-attention-based multi-modal architecture for emergency triage prediction. Rather than enforcing rigid spatial dependencies on triage data, the proposed model directly concatenates numerical and textual embeddings (via TabNet and BERT) along the channel dimension. By leveraging a Vision Transformer (ViT) encoder, this approach explicitly models the deep, cross-modal global interactions between vital signs and chief complaints, producing a unified, robust feature representation for accurate triage classification.

The remainder of this paper is organized as follows: Section \ref{sec2} describes the related works; Section \ref{sec3} provides a brief methodology of the original architecture and a detailed explanation of ViT Encoder implementation. Then, Section \ref{sec4} presents the experimental results, and finally, we provide some discussions and concluding remarks in Section \ref{sec5}.

\section{Related Work}\label{sec2}
There is an increasing interest in combining ED with AI for initial triage, primarily using supervised machine learning algorithms. Various research has been done, from processing data at the initial triage level to predicting patients with critical care needs, patients' triage level, or hospitalization using deep neural network \cite{chai2024advancing}, convolutional neural network \cite{arnaud_deep_2020} and temporal convolutional network \cite{sharafat_patientflownet_2021}.

Attention-based models have been adopted as a powerful tool to ease the workload in ED. Attention-based machine learning has demonstrated improved results on the task application. Addressing the challenge of patient drop-out during questionnaire-based digital triage, Krylova et al. \cite{krylova2024leveraging} investigated the efficacy of various machine learning models including attention-based TabTransformer in predicting patient urgency levels. Their findings show that the TabTransformer demonstrated robust performance, maintaining over 80\% accuracy across all tested degrees of interview completeness (from 100\% down to 40\%). However, textual clinical features were entirely omitted from the models, restricting the architecture to structured questionnaire data alone. On the other hand, deep learning methods such bi-LSTM \cite{gligorijevic_deep_2018} and BERT \cite{tahayori_advanced_2021} have been used to extract semantic features from textual chief complaints. The studies show that the proposed models achieved promising predictive performance in ED decision-making tasks. However, because the models are limited to textual chief complaints only, they cannot capture the relationships between text and numerical vital signs.

Chen et al. proposes a predictive framework that integrates advanced natural language processing with deep neural network for predicting hospital admission from multi-modal triage data \cite{chen2022imbalanced}. The textual chief complaints are converted into semantic feature vectors using bi-LSTM with an attention mechanism. Then, the textual embeddings are fused with numerical patient features such as vital signs and demographic to form a single feature vector. A similar study is reported in \cite{wang_deeptriager_2019}, whereby a deep learning model called DeepTriager that processes both structured vital signs and textual clinical records for emergency department triage. The architecture uses a representation module where a bi-LSTM and a self-attention layer extract contextual features from text, while parallel fully connected layers process structured data. These features are then integrated in a fusion module using either concatenation or a bilinear-combination approach. 

A multimodal deep learning model is proposed by K. C. Leung et al. \cite{leung_novel_2021}, which uses TabNet \cite{arik2021tabnet} to process vital signs data and BERT to process chief complaints. The extracted vital sign embeddings and text embeddings are reshaped and concatenated into an image-like tensor before being passed to a series of 2D convolutional and max-pooling layers for classification. Similar approach is presented in \cite{lin2024interpretable} whereby TabNet and a pretrained MacBERT are used to extract semantic features from heterogeneous electronic health records and fuses them into unified embeddings. These embeddings are then passed to a multi-task CNN and multilayer perceptron (MLP) classifiers to predict triage level and hospitalization. However, these existing models rely on convolutional and downsampling layers, restricting feature extraction to fixed, localized sliding windows. This feature learning pipeline is structurally incapable of capturing long-range, cross-modal global interactions between vital sign and textual embeddings. Furthermore, during the concatenation, the embeddings are reshaped into grid matrices to satisfy standard 2D convolutional operations, which imposes a rigid spatial dependency that does not align with heterogenous, non-grid electronic health record data. 

To address these limitations, our proposed model transforms the localized, convolutional fusion network into a global, self-attention-based network. Rather than compressing or reshaping the embeddings during fusion, our model directly concatenates the matrices along the channel dimension to form a multi-modal feature map before passing them to a Vision Transformer (ViT) encoder. By treating the combined modalities as a single multi-channel feature map, our architecture leverages ViT's self-attention mechanism to explicitly model the deep, global interactions between numerical and textual features, providing a more integrated and robust representation for precise triage prediction.

\section{Methodology}\label{sec3}
\subsection{Baseline Model}\label{subsec3_1}
This study aims to improve the baseline model proposed reported in \cite{leung_novel_2021} for triage zone classification. The baseline model employs two feature learning pipelines: a TabNet Encoder for vital signs and BERT for chief complaint text. The TabNet output ($1 \times 64$) and BERT embedding ($1 \times 768$) are reshaped into $1 \times 8 \times 8$ and $12 \times 8 \times 8$, respectively, and concatenated to form a $13 \times 8 \times 8$ feature map. This image-like representation is processed by CNN layers with max-pooling and ReLU activation, followed by fully connected layers for final prediction.

\subsection{Proposed Model}\label{subsec3_2}
The proposed model includes similar feature learning pipelines, TabNet Encoder and Bert for processing vital sign and chief complaint respectively. Unlike the baseline model, the proposed model exploits the Vision Transformer (ViT) to model the dependencies between the multimodal inputs. The ViT serves as the core architecture in our model, adeptly handling the multi-dimensional nature of triage data, which encompasses both numerical and textual information. This approach allows us to maintain the inherent spatial structure and relational context within the data. We streamline the computational process by forgoing the image patching step used for processing larger images. This ensures that our methodology is both computationally efficient and well-suited to the input data scale.

\subsubsection{Model Architecture}\label{subsubsec3_2_1}
\begin{figure*}[t]
\centering
\includegraphics[width=0.8\textwidth]{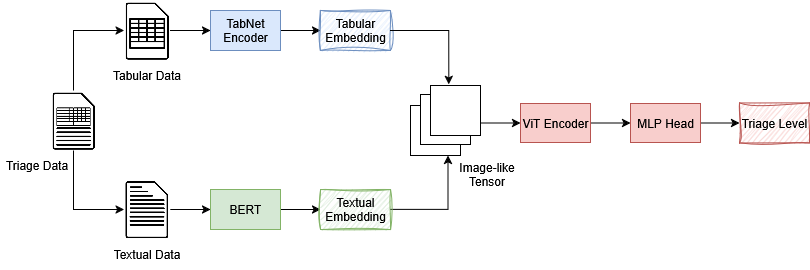}
\caption{Proposed model architecture}\label{fig:model_architecture}
\end{figure*}

The proposed model architecture processes tabular and textual triage data streams using TabNet, BERT, and a Vision Transformer (ViT) Encoder. As depicted in Figure \ref{fig:model_architecture}, the model architecture integrates integrates vital signs and chief complaints into a unified representation for triage zone classification. Tabular inputs including vital signs alongside other numerical and categorical features are processed by TabNet. To learn dataset-specific patterns without labeled data, the TabNet encoder employs unsupervised pre-training to reconstruct masked features. This optimization minimizes a reconstruction loss normalized by feature variance, where a binary mask $S \in {0,1}^{D_{tabular}}$ obscures 80\% of features (pre-training ratio = 0.8).

The pre-training uses a feature transformer with two shared and two step-dependent fully connected (FC) layers, incorporating batch normalization and gated linear unit (GLU) non-linearities. It also includes an attentive transformer that generates sparsemax-normalized masks $M[i] \in \mathbb{R}^{D_{tabular}}$, where the feature dimension is $D_{tabular}=32$. This design selects salient features at each step to enhance both efficiency and interpretability. 

For the supervised learning phase, the weights of the pre-trained TabNet encoder are loaded for fine-tuning. Given the raw tabular input, $x_{tabular} \in \mathbb{R}^{D_{tabular}}$, the encoder produces a tabular embedding $z_{TabNet} \in \mathbb{R}^{64}$. This embedding is the summed output across $n_{steps}=3$ decision steps, where each step yields a decision embedding $d[i] \in \mathbb{R}^{64}$. Then, $z_{TabNet}$ is reshaped to $f_{TabNet} \in \mathbb{R}^{1 \times 8 \times 8}$ (1 channel, $8 \times 8$ spatial dimensions) to align with the input format of the ViT.

In parallel, textual chief complaint is processed by a pre-trained BERT model bert-base-uncased via SentenceTransformer, which embeds the input $x_{text}$ into a contextualized embedding vector $z_{BERT} \in \mathbb{R}^{D_{BERT}}$, where $D_{BERT}=768$. The BERT model consists of 12 transformer encoder layers that employ self-attention to capture contextual relationships within the input text. The resulting embedding vector is then reshaped into $f_{BERT} \in R^{(12 \times 8 \times 8)}$, comprising 12 channels with spatial dimension $H=W=8$). Finally, the TabNet and BERT feature maps are concatenated along the channel dimension to produce the unified feature representation $f_{c} = [f_{TabNet},f_{BERT}] \in R^{(K+L) \times H \times W}$, where $K=1$ (TabNet channels), $L=12$ (BERT channels) and $H=W=8$.

The resulting $f_c \in \mathbb{R}^{(13 \times 8 \times 8)}$ contains 832 features when flattened. Unlike conventional ViT applications, where images are divided into patches, our concatenated feature map $f_c$, with only 832 features, is significantly smaller, making patching computationally unnecessary. The flattened sequence $x_{flat} \in \mathbb{R}^{832}$ is combined with sinusoidal 1D positional embeddings, generated using a sine-cosine function, to encode the spatial positions of features, enabling the ViT to model their relative positions, as in standard transformer models, which is essential for understanding triage data dependencies.

The ViT Encoder, configured with 8 layers, 128 multihead self-attention heads, and a feedforward network dimension of 2048, processes this sequence to model global interactions between numerical and textual features (e.g., linking a high heart rate to specific chief complaints). The output is fed to an MLP head, which maps the 1024-dimensional transformer output to three triage zones. This architecture leverages TabNet's pre-trained encoder for clinically relevant numerical feature prioritization, BERT's contextual text processing, and ViT's global modeling to integrate both modalities into a cohesive triage prediction model.

\subsubsection{Implementation and Model Training}\label{subsubsec3_2_2}
Both the baseline and proposed models were optimized using Optuna for hyperparameter tuning. The optimizer (Adam or AdamW) and learning rate (1e-9 to 1, logarithmic scale) were optimized through 50 trials. Experiments were carried out on a workstation equipped with AMD Ryzen 7 6800H 4.7GHz, 16GB memory, and Nvidia GeForce RTX 3070 Ti graphic cards with 8GB virtual memory. The optimal parameters obtained were AdamW with a learning rate of 0.003 for the baseline model and Adam with a learning rate of 0.01 for the proposed model.

The datasets were stratified into training (80\%), validation (10\%), and test (10\%) sets. Models were trained for up to 500 epochs with early stopping after 30 epochs without improvement. The model weights with the lowest weighted validation loss were saved and used for testing. F1-score and accuracy were evaluated on the test set, and the experiment was repeated 10 times to obtain average performance metrics. On average, across 10 experiments, the baseline model triggered early stopping at 17.8 epochs with a validation loss of 0.501, showing rapid overfitting while our model averaged 63.2 epochs before early stopping, achieving a lower validation loss of 0.474. This superior generalization is primarily driven by the inclusion of attention mechanisms and layer normalization. Appendix \ref{appendix_loss_curves} shows an example of the training and validation losses of the baseline model and the proposed model. 

\section{Experimental Setup and Results}\label{sec4}
\subsection{Dataset Description}\label{sec4_1}
The triage dataset was collected from emergency department of Hospital Universiti Sains Malaysia (HUSM). Institutional ethics committee approval was obtained prior to data collection in accordance with the research ethics committee of Universiti Sains Malaysia (USM/JEPeM/20090486). The data collected for this study are during 2017 until 2019, we exclude recent data from 2020 and above because Covid might cause unprecedented changes. Overall, a total of 11,102 triage data were collected for this study with 592 (5\%) in red zone, 4272 (39\%) in yellow zone and 6,236 (56\%) in green zone. Due to the clinical sensitivity of the triage data, we avoided synthetic oversampling techniques like SMOTE to preserve data integrity. Instead, we addressed class imbalance during training by implementing a weighted cross-entropy loss function, assigning higher weights to the underrepresented red zone class (inversely proportional to class frequencies). Publicly available triage data contain fewer features than our data, making it impossible to merge or utilize the data with our own data. 

\subsection{Data Preprocessing}\label{sec4_2}
The raw triage data which consists of 25 columns is pre-processed so that it can be used for modeling. For each patient, the age is derived from the date of birth (DOB) and the registration date and time. For example, a patient born on 1st January 2009, who registered on 1st March 2019, the calculated age is 10.1667. Age is calculated in days only for infants under one month old. Finally, we normalize age data to a scale between 0 and 150.

Next, we pre-process heart rate, respiratory rate, oxygen saturation, temperature and blood pressure by checking for typos. If any error is found, we either correct them or let them as missing value. We also refined blood pressure data from 92/64 to 92 as systole and 64 as diastole. We continued the process by assigning threshold levels based on the patients' age and their data. For example, an adult patient with a heart rate lower than 60 will be assigned as bradycardia. We assign 1 for when data reach or pass certain thresholds and 0 when the data is in normal threshold. The threshold for each data is based on the data given from doctors in HUSM. Then, those data were normalized with heart rate in range of 0 to 250, respiratory rate in range of 0 to 80, oxygen saturation in range of 0 to 100, temperature in range of 20 to 45, systole in range of 0 to 250 and diastole in range of 0 to 200.

After that, we one-hot encode categorical data, which includes patients' signs, patients' general conditions, most occurring comorbidities and arrival mode. Patients' signs consist of Traumatic, Non-Traumatic, Medico-Legal and Non-Medico-Legal. Any missing values will be replaced with Non-Traumatic and Non-Medico-Legal. General conditions include Alert, Drowsy, Unresponsive, Breathlessness, Pale and Weak, and any missing values will be replaced with Alert. Most occurring comorbidities are Hypertension, Diabetes, Heart Disease, Asthma, Other Comorbid and No Comorbid. Arrival mode consists of Own Transport, Ambulance, Police and Others, any missing values will be replaced with Own Transport. Then, we assign a binary for a patient's pregnancy by assigning 1 if the patient is pregnant and 0 if the patient is not pregnant.

We proceed by pre-processing the chief complaint which is in the form of text data. Chief complaints are the statement from patients describing their symptoms, problems or other reasons for their visit to ED. This text data also sometimes contains short forms. Triagers usually use short forms to save time, so they wrote a shortened form that represents the full word. Therefore, the short forms are expanded to make them human readable and processable by BERT. All texts are lowercased to ensure consistent text representation, and leading and trailing whitespaces are removed. An example of chief complaint pre-processing can be seen in Table \ref{tab:text-preprocessing}.

The triage zone is encoded as 0 (green), 1 (yellow), and 2 (red). Missing values are imputed using Multivariate Imputation by Chained Equations (MICE) with the Bayesian Ridge algorithm. The final dataset consists of 32 structured features, one pre-processed text feature, and one target label column. The whole completed pre-processed data is provided in Appendix \ref{appendix_preprocessed_data}.

\begin{table*}[tb]
\centering
\caption{Example of text pre-processing}
\label{tab:text-preprocessing}
\begin{tabularx}{0.9\textwidth}{l X}
\toprule
\textbf{Stage} & \textbf{Text Sample} \\
\midrule
Before pre-processing & mva mb vs car, loc, l/w @ Lt u/l \\
\addlinespace
After pre-processing & motor vehicle accident motor bike vs car and loss of consciousness and laceration wound at left upper limb \\
\bottomrule
\end{tabularx}
\end{table*}

\subsection{Results and Discussion}\label{sec4_3}
For the evaluation, we used macro average instead of weighted average because weighted average considers the number of actual occurrences of the class in dataset, this will put more weight on green zone. The macro ROC AUC was calculated by Averaging the area under the ROC curve for each class. Our model outperformed the baseline across the 10 experiments in F1-score, and fell short only once in accuracy and ROC AUC, achieving higher overall averages (Accuracy: 80.19\% vs. 78.66\%; F1-score: 70.09\% vs. 68.39\%; ROC AUC: 90.59\% vs. 89.33\%) as shown in Table \ref{tab:performance_comparison}. Appendix \ref{appendix_performance_comparison} lists the classification performances for all 10 experiments. 

\begin{table}[h]
\centering
\caption{Performance comparison of baseline and proposed models}
\begin{tabular}{lcc}
\hline
\textbf{Metric} & \textbf{Baseline} & \textbf{Our Model} \\
\hline
Average Accuracy & 78.66 & 80.19 \\
Average F1-score & 68.39 & 70.09 \\
Average ROC AUC & 89.33 & 90.59 \\
\hline
\end{tabular}
\label{tab:performance_comparison}
\end{table}

\begin{figure}[t]
\centering
\includegraphics[width=0.49\textwidth]{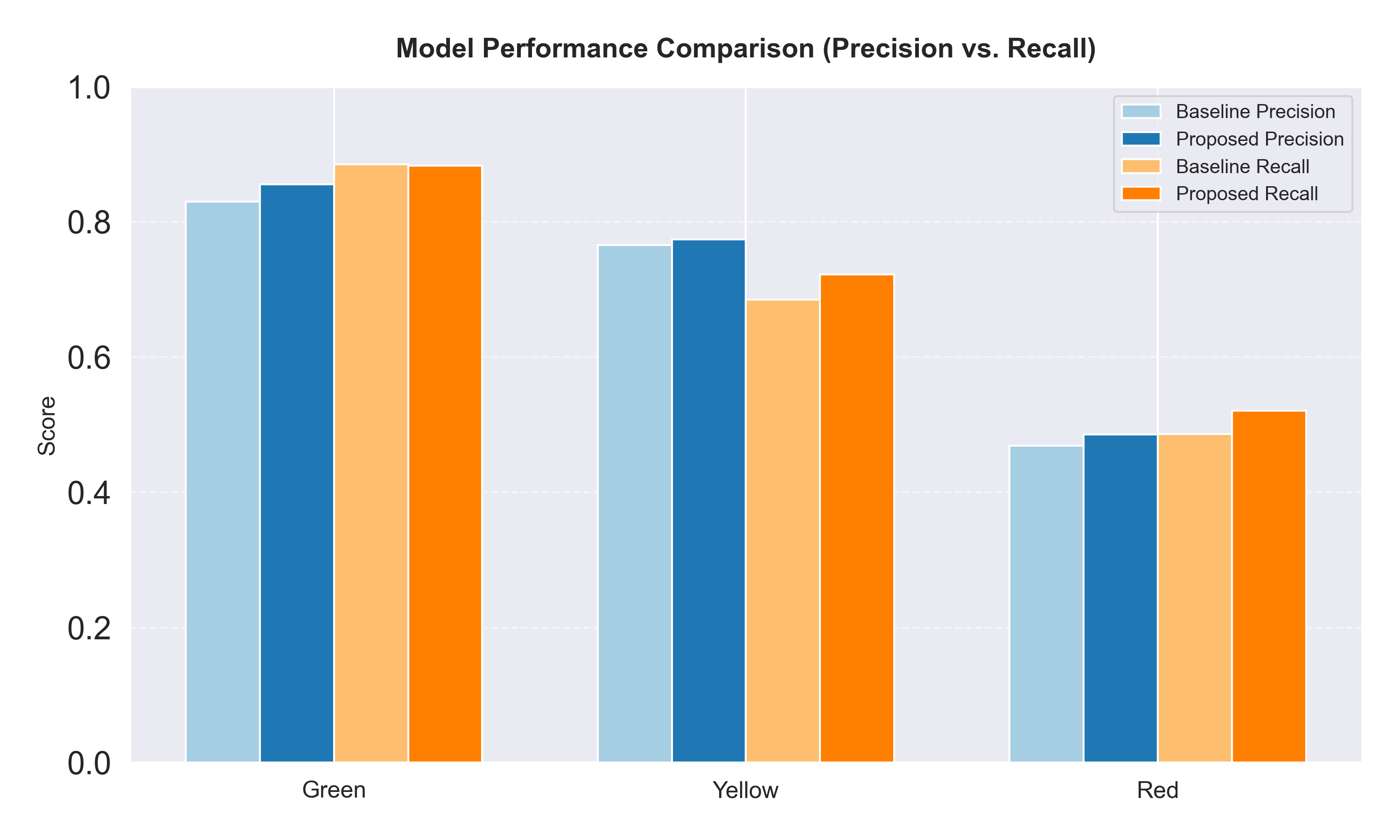}
\caption{Proposed model architecture}\label{fig:recall_precision}
\end{figure}

Two-tailed paired t-tests across 10 splits confirmed these improvements are statistically significant ($p < 0.05$). Accuracy gained 1.947\% ($t=3.675, p=0.005$), F1-score gained 2.518\% ($t=4.107, p=0.003$), and ROC AUC gained 1.308\% ($t=4.404, p=0.002$). These relative gains demonstrate enhanced triage classification, particularly for red and yellow zones, potentially reducing under-triage by 2-4\% and improving patient outcomes in high-pressure emergency departments. Figure \ref{fig:recall_precision} compares the average class precision and recall across the 10 experiments. While our proposed model outperformed the baseline in all precision measures, it only achieved higher recall in the yellow and red zones. For the green class, the baseline model maintained a slightly higher average recall of 0.885 compared to our model's 0.883.

\begin{figure}[htbp]
    \centering
    \begin{subfigure}[b]{0.35\textwidth}
        \centering
        \includegraphics[width=\textwidth]{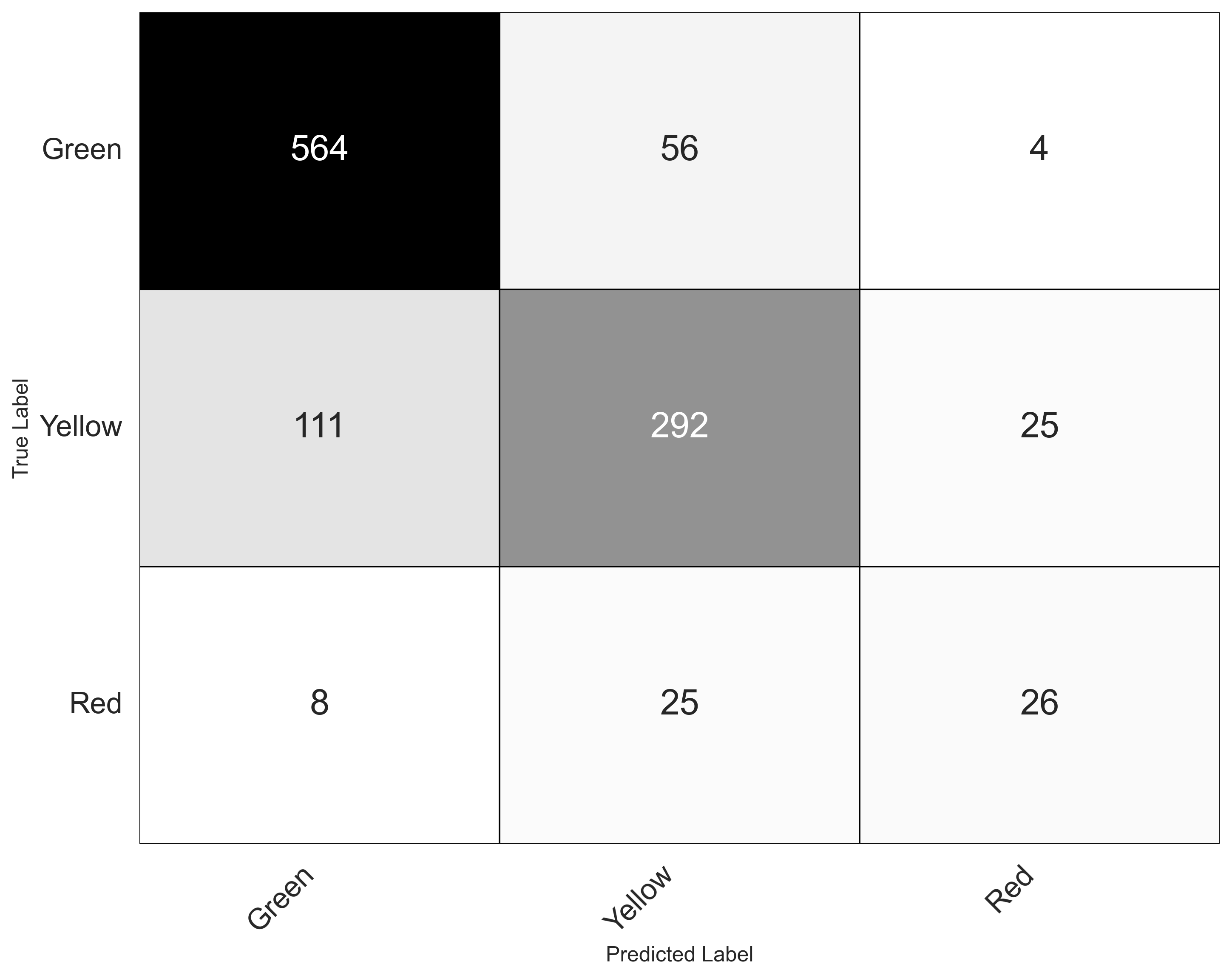} 
        \caption{}
        \label{fig:sub-first}
    \end{subfigure}
    \hfill 
    \begin{subfigure}[b]{0.35\textwidth}
        \centering
        \includegraphics[width=\textwidth]{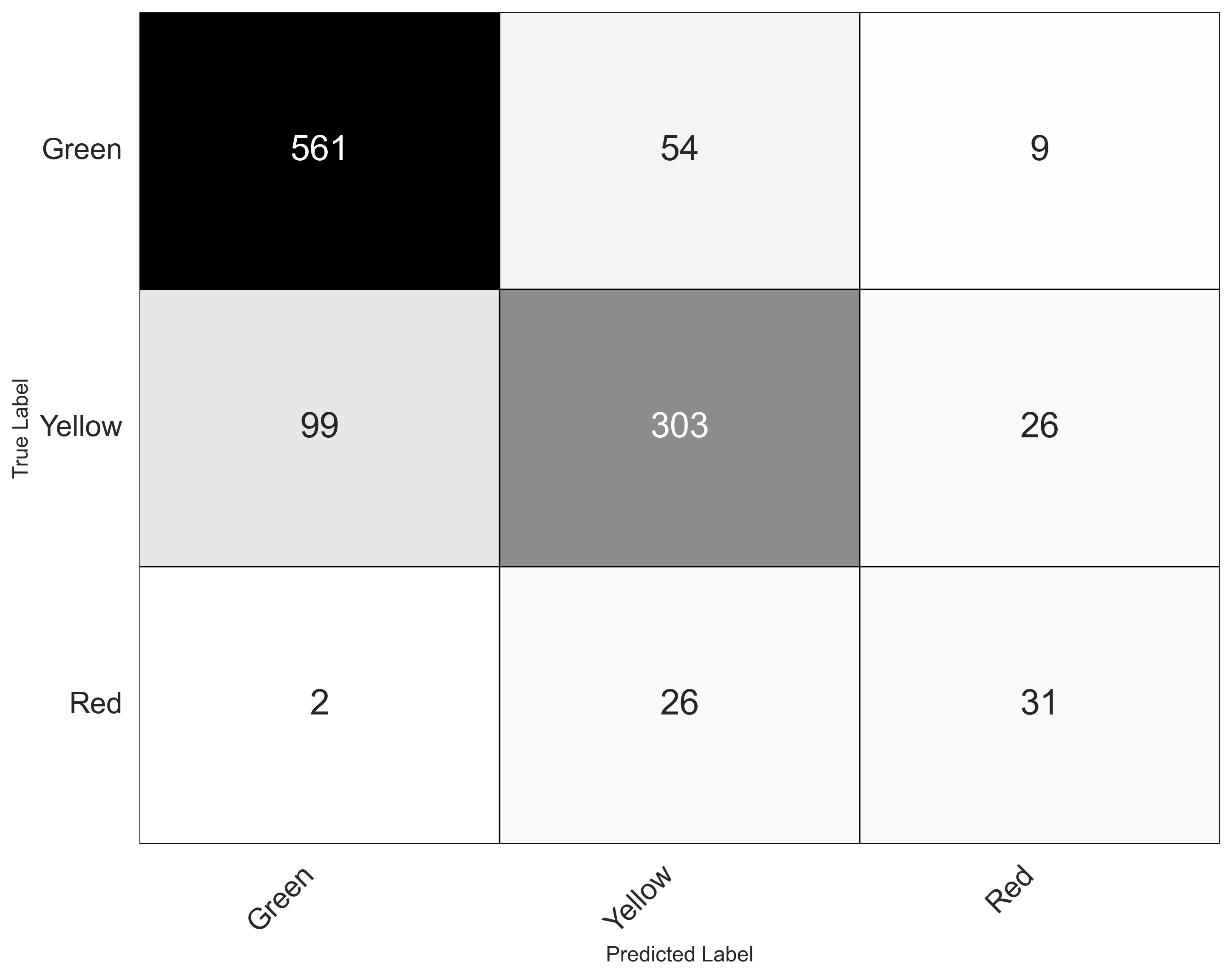} 
        \caption{}
        \label{fig:sub-second}
    \end{subfigure}
    
    \caption{Confusion matrix of (a) baseline model (b) proposed model.}
    \label{fig:baseline_model_conf_matrix}
\end{figure}

Figure \ref{fig:baseline_model_conf_matrix} compares the confusion matrices from the fifth experiment. Across all 10 experiments, the proposed model matched the baseline in Red zone classification and consistently outperformed it in classifying the Yellow zone. For the Green zone, the baseline performed slightly better, achieving higher True Positive counts in give experiments compared to the proposed model's four, with one experiment resulting in a tie. 

An error analysis across all 10 runs (~13,876 test cases) reveals specific misclassification patterns. For the critical Red zone, the proposed model minimized severe under-triage (R $\rightarrow$ G), dropping to 66 errors compared to the baseline's 74. However, it misclassified slightly more Red zone patients as Yellow (R $\rightarrow$ Y: 319 vs. 306). Consequently, total Red zone under-triages were comparable, with the proposed model totaling 385 versus the baseline's 380.

In the Yellow zone, the proposed model demonstrated a significant advantage by substantially reducing under-triage (Y $\rightarrow$ G), recording 1,157 errors against the baseline's 1,375. It also minimized over-triage (Y $\rightarrow$ R: 333 vs. 402). This simultaneous reduction prevents urgent patients from being safely de-prioritized while optimizing hospital resource allocation.

For Green zone over-triage, both models exhibited nearly identical performance. Misclassifications to the Yellow Zone (G $\rightarrow$ Y) were highly similar, with 864 instances for the proposed model and 858 for the baseline. Extreme over-triage (G $\rightarrow$ R) was identical, with each model recording exactly 60 instances across the 10 experiments.

In summary, while the proposed model's Red Zone accuracy is comparable to the baseline, it offers distinct advantages. The proposed model reduces high-risk Red-to-Green under-triage and significantly improves Yellow Zone management by decreasing both under-triage (to Green) and over-triage (to Red). These findings suggest the proposed model can enhance patient safety and resource prioritization, particularly for the large Yellow Zone cohort.

\subsection{Comparison with State-of-the-art Models}\label{sec4_4}
We compare our model against state-of-the-art multi-modal triage models from Section \ref{sec2}. Table \ref{tab:relevant-work} summarizes the existing model and their performances. Compared to the baseline model, our model achieves 80.19\% accuracy and 70.09\% F1-score, outperforming the baseline by 1.95\% and 2.49\%. Another study with a similar approach to the baseline model reported accuracies of 0.633 and 0.822 for triage and hospitalization classification, respectively \cite{lin2024interpretable}. Gligorijevic et al. \cite{gligorijevic_deep_2018} achieved 79.21\% accuracy using bi-LSTM with attention but represented vital signs as word vectors, which may introduce semantic mismatches. Our approach improves robustness by effectively integrating numerical features.

BERT-based triage prediction was used to predict ED patient disposition as either ``Home'' or ``Admit to Ward'' \cite{tahayori_advanced_2021}. The model achieved 83\% accuracy and 63\% F1-score after text preprocessing and oversampling. However, it excludes vital signs, potentially missing important clinical conditions. DeepTriager predicts ED acuity using structured and textual data with bi-LSTM and self-attention \cite{wang_deeptriager_2019}. It achieved average AUC scores of 0.9594 using bilinear fusion and 0.956 using concatenation. However, its fully connected layers treat structured features independently, limiting feature dependency learning. Additionally, it only considers a limited set of vital signs.

Chen et al. \cite{chen_imbalanced_2022} proposed an attention-based model for imbalanced emergency triage data. Their bi-LSTM with Attention Mechanism achieved 95\% accuracy and 92\% F1-score, improving to 96\% with SMOTE. Despite strong performance, the model only uses textual data and does not incorporate vital signs. Our proposed model addresses these limitations by incorporating TabNet for contextual vital sign representation and ViT with attention heads to capture global feature relationships, unlike conventional CNN-based multimodal approaches such as \cite{leung_novel_2021} and \cite{lin2024interpretable}.

\begin{table*}[t!]
\centering
\small
\caption{Comparison of relevant works and the proposed model}
\label{tab:relevant-work}
\begin{tabularx}{\textwidth}{l X X}
\toprule
\textbf{Relevant Work} & \textbf{Description} & \textbf{Performance} \\
\midrule
S. Krylova et al. \cite{krylova2024leveraging} & Utilize TabTransformer to predict Triage24 data of unfinished questionnaire-based digital medical triage interviews. & 80\% accuracy \\
\addlinespace
D. Gligorijevic et al. \cite{gligorijevic_deep_2018} & Predicts ED resources needed using a text embedding block, bi-LSTM, attention block, and classification block. & 79.21\% accuracy, 0.8797 AUROC \\
\addlinespace
B. Tahayori et al. \cite{tahayori_advanced_2021} & Predicts patient disposition in ED from triage notes using BERT, with pre-processing and data balancing techniques. & 83\% accuracy, 63\% F1-score \\
\addlinespace
T. L. Chen et al. \cite{chen_imbalanced_2022} & Addresses class imbalance in ED data using bi-LSTM with Attention Mechanism and SMOTE for data preprocessing. & 95\% accuracy, 92\% F1-score (96\% with SMOTE) \\
\addlinespace
G. Wang et al. \cite{wang_deeptriager_2019} & Predicts ED patient acuity level using structured and textual data, with bi-LSTM, self-attention, and feature fusion modules. & 0.9594 AUC (bilinear-combination), 0.956 AUC (concatenation) \\
\addlinespace
Y. T. Lin et al. \cite{lin2024interpretable} & Utilize TabNet and MacBert with convolutional layers for triage and hospitalization prediction & 0.633 accuracy (triage classification), 0.822 accuracy (hospitalization) \\
\addlinespace
K. C. Leung et al. \cite{leung_novel_2021} & Utilize TabNet and Bert with convolutional layers for hospitalization prediction & 0.805 accuracy, 0.836 AUC \\
\bottomrule
\end{tabularx}
\end{table*}

\section{Conclusion}\label{sec5}
This study introduces a novel multimodal deep learning model for emergency triage zone prediction, integrating TabNet for numerical vital signs and BERT for textual chief complaints, with a Vision Transformer (ViT) Encoder to capture global feature relationships. The proposed model fuses the vital sign and textual embeddings into a unified feature map, which is subsequently processed by the ViT Encoder to learn cross-modal dependencies between physiological and textual information. This architecture enables the model to jointly exploit structured and unstructured clinical data for improved triage decision-making. The proposed model is evaluated on 11,102 HUSM triage records, achieving a significant immprovement over the baseline model by +1.95\% (accuracy) and +2.49\% (F1-score). Future work will explore multi-center datasets to enhance generalizability and alternative transformer architectures to reduce computational demands, ensuring scalability in diverse clinical settings.

\backmatter

\bmhead{Acknowledgements}
We would like to express our sincere gratitude to Universiti Sains Malaysia for their generous support through the Research University (RUI) Grant (1001/PPSP/8014125).

\section*{Declarations}

\begin{itemize}
\item Funding: This work was supported by Universiti Sains Malaysia through Research University (RUI) Grant (1001/PPSP/8014125)
\item Conflict of interest: The authors declare that they have no known competing financial interests or personal relationships that could have appeared to influence the work reported in this paper.
\item Ethics approval and consent to participate: Institutional ethics committee approval was obtained prior to data collection in accordance with the research ethics committee of Universiti Sains Malaysia (USM/JEPeM/20090486).
\item Code availability: Source code is available at https://github.com/Hazqeel09/Multimodal-Attention-based-Deep-Learning.
\item Data availability: The datasets generated and/or analysed during the current study are available in the UCI Machine Learning repository, https://archive.ics.uci.edu/dataset/1312/emergency+department+multimodal+patient+triage+(hospital+usm). [Submitted pending approval]
\end{itemize}

\bibliography{triage_bibliography}


\begin{thebibliography}{23}
\ifx \bisbn   \undefined \def \bisbn  #1{ISBN #1}\fi
\ifx \binits  \undefined \def \binits#1{#1}\fi
\ifx \bauthor  \undefined \def \bauthor#1{#1}\fi
\ifx \batitle  \undefined \def \batitle#1{#1}\fi
\ifx \bjtitle  \undefined \def \bjtitle#1{#1}\fi
\ifx \bvolume  \undefined \def \bvolume#1{\textbf{#1}}\fi
\ifx \byear  \undefined \def \byear#1{#1}\fi
\ifx \bissue  \undefined \def \bissue#1{#1}\fi
\ifx \bfpage  \undefined \def \bfpage#1{#1}\fi
\ifx \blpage  \undefined \def \blpage #1{#1}\fi
\ifx \burl  \undefined \def \burl#1{\textsf{#1}}\fi
\ifx \doiurl  \undefined \def \doiurl#1{\url{https://doi.org/#1}}\fi
\ifx \betal  \undefined \def \betal{\textit{et al.}}\fi
\ifx \binstitute  \undefined \def \binstitute#1{#1}\fi
\ifx \binstitutionaled  \undefined \def \binstitutionaled#1{#1}\fi
\ifx \bctitle  \undefined \def \bctitle#1{#1}\fi
\ifx \beditor  \undefined \def \beditor#1{#1}\fi
\ifx \bpublisher  \undefined \def \bpublisher#1{#1}\fi
\ifx \bbtitle  \undefined \def \bbtitle#1{#1}\fi
\ifx \bedition  \undefined \def \bedition#1{#1}\fi
\ifx \bseriesno  \undefined \def \bseriesno#1{#1}\fi
\ifx \blocation  \undefined \def \blocation#1{#1}\fi
\ifx \bsertitle  \undefined \def \bsertitle#1{#1}\fi
\ifx \bsnm \undefined \def \bsnm#1{#1}\fi
\ifx \bsuffix \undefined \def \bsuffix#1{#1}\fi
\ifx \bparticle \undefined \def \bparticle#1{#1}\fi
\ifx \barticle \undefined \def \barticle#1{#1}\fi
\bibcommenthead
\ifx \bconfdate \undefined \def \bconfdate #1{#1}\fi
\ifx \botherref \undefined \def \botherref #1{#1}\fi
\ifx \url \undefined \def \url#1{\textsf{#1}}\fi
\ifx \bchapter \undefined \def \bchapter#1{#1}\fi
\ifx \bbook \undefined \def \bbook#1{#1}\fi
\ifx \bcomment \undefined \def \bcomment#1{#1}\fi
\ifx \oauthor \undefined \def \oauthor#1{#1}\fi
\ifx \citeauthoryear \undefined \def \citeauthoryear#1{#1}\fi
\ifx \endbibitem  \undefined \def \endbibitem {}\fi
\ifx \bconflocation  \undefined \def \bconflocation#1{#1}\fi
\ifx \arxivurl  \undefined \def \arxivurl#1{\textsf{#1}}\fi
\csname PreBibitemsHook\endcsname

\bibitem[\protect\citeauthoryear{Fry and Burr}{2002}]{fry_review_2002}
\begin{barticle}
\bauthor{\bsnm{Fry}, \binits{M.}},
\bauthor{\bsnm{Burr}, \binits{G.}}:
\batitle{Review of the triage literature: {Past}, present, future?}
\bjtitle{Australian Emergency Nursing Journal}
\bvolume{5}(\bissue{2}),
\bfpage{33}--\blpage{38}
(\byear{2002})
\doiurl{10.1016/S1328-2743(02)80018-9}
\end{barticle}
\endbibitem

\bibitem[\protect\citeauthoryear{Joseph et~al.}{2023}]{joseph_use_2023}
\begin{barticle}
\bauthor{\bsnm{Joseph}, \binits{M.J.}},
\bauthor{\bsnm{Summerscales}, \binits{M.}},
\bauthor{\bsnm{Yogesan}, \binits{S.}},
\bauthor{\bsnm{Bell}, \binits{A.}},
\bauthor{\bsnm{Genevieve}, \binits{M.}},
\bauthor{\bsnm{Kanagasingam}, \binits{Y.}}:
\batitle{The use of kiosks to improve triage efficiency in the emergency
  department}.
\bjtitle{npj Digital Medicine}
\bvolume{6}(\bissue{1}),
\bfpage{1}--\blpage{9}
(\byear{2023})
\doiurl{10.1038/s41746-023-00758-2}
\end{barticle}
\endbibitem

\bibitem[\protect\citeauthoryear{Tam et~al.}{2018}]{tam_review_2018}
\begin{barticle}
\bauthor{\bsnm{Tam}, \binits{H.L.}},
\bauthor{\bsnm{Chung}, \binits{S.F.}},
\bauthor{\bsnm{Lou}, \binits{C.K.}}:
\batitle{A review of triage accuracy and future direction}.
\bjtitle{BMC Emergency Medicine}
\bvolume{18}(\bissue{1}),
\bfpage{58}
(\byear{2018})
\doiurl{10.1186/s12873-018-0215-0}
\end{barticle}
\endbibitem

\bibitem[\protect\citeauthoryear{Kumar et~al.}{2023}]{kumar_artificial_2023}
\begin{barticle}
\bauthor{\bsnm{Kumar}, \binits{Y.}},
\bauthor{\bsnm{Koul}, \binits{A.}},
\bauthor{\bsnm{Singla}, \binits{R.}},
\bauthor{\bsnm{Ijaz}, \binits{M.F.}}:
\batitle{Artificial intelligence in disease diagnosis: a systematic literature
  review, synthesizing framework and future research agenda}.
\bjtitle{Journal of Ambient Intelligence and Humanized Computing}
\bvolume{14}(\bissue{7}),
\bfpage{8459}--\blpage{8486}
(\byear{2023})
\doiurl{10.1007/s12652-021-03612-z}
\end{barticle}
\endbibitem

\bibitem[\protect\citeauthoryear{Javaid
  et~al.}{2022}]{javaid_significance_2022}
\begin{barticle}
\bauthor{\bsnm{Javaid}, \binits{M.}},
\bauthor{\bsnm{Haleem}, \binits{A.}},
\bauthor{\bsnm{Pratap~Singh}, \binits{R.}},
\bauthor{\bsnm{Suman}, \binits{R.}},
\bauthor{\bsnm{Rab}, \binits{S.}}:
\batitle{Significance of machine learning in healthcare: {Features}, pillars
  and applications}.
\bjtitle{International Journal of Intelligent Networks}
\bvolume{3},
\bfpage{58}--\blpage{73}
(\byear{2022})
\doiurl{10.1016/j.ijin.2022.05.002}
\end{barticle}
\endbibitem

\bibitem[\protect\citeauthoryear{Davenport and
  Kalakota}{2019}]{davenport_potential_2019}
\begin{barticle}
\bauthor{\bsnm{Davenport}, \binits{T.}},
\bauthor{\bsnm{Kalakota}, \binits{R.}}:
\batitle{The potential for artificial intelligence in healthcare}.
\bjtitle{Future Healthcare Journal}
\bvolume{6}(\bissue{2}),
\bfpage{94}--\blpage{98}
(\byear{2019})
\doiurl{10.7861/futurehosp.6-2-94}
\end{barticle}
\endbibitem

\bibitem[\protect\citeauthoryear{Raita et~al.}{2019}]{raita_emergency_2019}
\begin{barticle}
\bauthor{\bsnm{Raita}, \binits{Y.}},
\bauthor{\bsnm{Goto}, \binits{T.}},
\bauthor{\bsnm{Faridi}, \binits{M.K.}},
\bauthor{\bsnm{Brown}, \binits{D.F.M.}},
\bauthor{\bsnm{Camargo}, \binits{C.A.}},
\bauthor{\bsnm{Hasegawa}, \binits{K.}}:
\batitle{Emergency department triage prediction of clinical outcomes using
  machine learning models}.
\bjtitle{Critical Care (London, England)}
\bvolume{23}(\bissue{1}),
\bfpage{64}
(\byear{2019})
\doiurl{10.1186/s13054-019-2351-7}
\end{barticle}
\endbibitem

\bibitem[\protect\citeauthoryear{Păvăloaia and
  Necula}{2023}]{pavaloaia_artificial_2023}
\begin{barticle}
\bauthor{\bsnm{Păvăloaia}, \binits{V.-D.}},
\bauthor{\bsnm{Necula}, \binits{S.-C.}}:
\batitle{Artificial {Intelligence} as a {Disruptive} {Technology}—{A}
  {Systematic} {Literature} {Review}}.
\bjtitle{Electronics}
\bvolume{12}(\bissue{5}),
\bfpage{1102}
(\byear{2023})
\doiurl{10.3390/electronics12051102}
\end{barticle}
\endbibitem

\bibitem[\protect\citeauthoryear{Mueller
  et~al.}{2022}]{mueller_artificial_2022}
\begin{barticle}
\bauthor{\bsnm{Mueller}, \binits{B.}},
\bauthor{\bsnm{Kinoshita}, \binits{T.}},
\bauthor{\bsnm{Peebles}, \binits{A.}},
\bauthor{\bsnm{Graber}, \binits{M.A.}},
\bauthor{\bsnm{Lee}, \binits{S.}}:
\batitle{Artificial intelligence and machine learning in emergency medicine: a
  narrative review}.
\bjtitle{Acute Medicine \& Surgery}
\bvolume{9}(\bissue{1}),
\bfpage{740}
(\byear{2022})
\doiurl{10.1002/ams2.740}
\end{barticle}
\endbibitem

\bibitem[\protect\citeauthoryear{Maslej-Krešňáková
  et~al.}{2020}]{maslej-kresnakova_comparison_2020}
\begin{barticle}
\bauthor{\bsnm{Maslej-Krešňáková}, \binits{V.}},
\bauthor{\bsnm{Sarnovský}, \binits{M.}},
\bauthor{\bsnm{Butka}, \binits{P.}},
\bauthor{\bsnm{Machová}, \binits{K.}}:
\batitle{Comparison of {Deep} {Learning} {Models} and {Various} {Text}
  {Pre}-{Processing} {Techniques} for the {Toxic} {Comments} {Classification}}.
\bjtitle{Applied Sciences}
\bvolume{10}(\bissue{23}),
\bfpage{8631}
(\byear{2020})
\doiurl{10.3390/app10238631} .
\bcomment{Number: 23}
\end{barticle}
\endbibitem

\bibitem[\protect\citeauthoryear{Leung et~al.}{2021}]{leung_novel_2021}
\begin{bchapter}
\bauthor{\bsnm{Leung}, \binits{K.-C.}},
\bauthor{\bsnm{Lin}, \binits{Y.-T.}},
\bauthor{\bsnm{Hong}, \binits{D.-Y.}},
\bauthor{\bsnm{Tsai}, \binits{C.-L.}},
\bauthor{\bsnm{Huang}, \binits{C.-H.}},
\bauthor{\bsnm{Fu}, \binits{L.-C.}}:
\bctitle{A {Novel} {Interpretable} {Deep}-{Learning}-{Based} {System} for
  {Triage} {Prediction} in the {Emergency} {Department}: {A} {Prospective}
  {Study}}.
In: \bbtitle{2021 {IEEE} {International} {Conference} on {Systems}, {Man}, and
  {Cybernetics} ({SMC})},
pp. \bfpage{2979}--\blpage{2985}
(\byear{2021}).
\doiurl{10.1109/SMC52423.2021.9658729}
\end{bchapter}
\endbibitem

\bibitem[\protect\citeauthoryear{Lin et~al.}{2024}]{lin2024interpretable}
\begin{barticle}
\bauthor{\bsnm{Lin}, \binits{Y.-T.}},
\bauthor{\bsnm{Deng}, \binits{Y.-X.}},
\bauthor{\bsnm{Tsai}, \binits{C.-L.}},
\bauthor{\bsnm{Huang}, \binits{C.-H.}},
\bauthor{\bsnm{Fu}, \binits{L.-C.}}:
\batitle{Interpretable deep learning system for identifying critical patients
  through the prediction of triage level, hospitalization, and length of stay:
  Prospective study}.
\bjtitle{JMIR Medical Informatics}
\bvolume{12},
\bfpage{48862}
(\byear{2024})
\end{barticle}
\endbibitem

\bibitem[\protect\citeauthoryear{Arik and Pfister}{2020}]{arik_tabnet_2020}
\begin{botherref}
\oauthor{\bsnm{Arik}, \binits{S.O.}},
\oauthor{\bsnm{Pfister}, \binits{T.}}:
{TabNet}: {Attentive} {Interpretable} {Tabular} {Learning}.
arXiv
(2020).
\doiurl{10.48550/arXiv.1908.07442} .
\url{http://arxiv.org/abs/1908.07442}
Accessed 2023-12-11
\end{botherref}
\endbibitem

\bibitem[\protect\citeauthoryear{Chai et~al.}{2024}]{chai2024advancing}
\begin{barticle}
\bauthor{\bsnm{Chai}, \binits{C.}},
\bauthor{\bsnm{Peng}, \binits{S.-z.}},
\bauthor{\bsnm{Zhang}, \binits{R.}},
\bauthor{\bsnm{Li}, \binits{C.-w.}},
\bauthor{\bsnm{Zhao}, \binits{Y.}}:
\batitle{Advancing emergency department triage prediction with machine learning
  to optimize triage for abdominal pain surgery patients}.
\bjtitle{Surgical Innovation}
\bvolume{31}(\bissue{6}),
\bfpage{583}--\blpage{597}
(\byear{2024})
\end{barticle}
\endbibitem

\bibitem[\protect\citeauthoryear{Arnaud et~al.}{2020}]{arnaud_deep_2020}
\begin{bchapter}
\bauthor{\bsnm{Arnaud}, \binits{{\'{E}}.}},
\bauthor{\bsnm{Elbattah}, \binits{M.}},
\bauthor{\bsnm{Gignon}, \binits{M.}},
\bauthor{\bsnm{Dequen}, \binits{G.}}:
\bctitle{Deep {Learning} to {Predict} {Hospitalization} at {Triage}:
  {Integration} of {Structured} {Data} and {Unstructured} {Text}}.
In: \bbtitle{2020 {IEEE} {International} {Conference} on {Big} {Data} ({Big}
  {Data})},
pp. \bfpage{4836}--\blpage{4841}
(\byear{2020}).
\doiurl{10.1109/BigData50022.2020.9378073} .
\burl{https://ieeexplore.ieee.org/document/9378073}
Accessed 2023-12-11
\end{bchapter}
\endbibitem

\bibitem[\protect\citeauthoryear{Sharafat and
  Bayati}{2021}]{sharafat_patientflownet_2021}
\begin{barticle}
\bauthor{\bsnm{Sharafat}, \binits{A.R.}},
\bauthor{\bsnm{Bayati}, \binits{M.}}:
\batitle{{PatientFlowNet}: {A} {Deep} {Learning} {Approach} to {Patient} {Flow}
  {Prediction} in {Emergency} {Departments}}.
\bjtitle{IEEE Access}
\bvolume{9},
\bfpage{45552}--\blpage{45561}
(\byear{2021})
\doiurl{10.1109/ACCESS.2021.3066164} .
Accessed 2023-12-11
\end{barticle}
\endbibitem

\bibitem[\protect\citeauthoryear{Krylova et~al.}{2024}]{krylova2024leveraging}
\begin{bchapter}
\bauthor{\bsnm{Krylova}, \binits{S.}},
\bauthor{\bsnm{Schmidt}, \binits{F.}},
\bauthor{\bsnm{Vlassov}, \binits{V.}}:
\bctitle{Leveraging machine learning models to predict the outcome of digital
  medical triage interviews}.
In: \bbtitle{2024 International Conference on Machine Learning and Applications
  (ICMLA)},
pp. \bfpage{160}--\blpage{167}
(\byear{2024}).
\bcomment{IEEE}
\end{bchapter}
\endbibitem

\bibitem[\protect\citeauthoryear{Gligorijevic
  et~al.}{2018}]{gligorijevic_deep_2018}
\begin{botherref}
\oauthor{\bsnm{Gligorijevic}, \binits{D.}},
\oauthor{\bsnm{Stojanovic}, \binits{J.}},
\oauthor{\bsnm{Satz}, \binits{W.}},
\oauthor{\bsnm{Stojkovic}, \binits{I.}},
\oauthor{\bsnm{Schreyer}, \binits{K.}},
\oauthor{\bsnm{Del~Portal}, \binits{D.}},
\oauthor{\bsnm{Obradovic}, \binits{Z.}}:
Deep {Attention} {Model} for {Triage} of {Emergency} {Department} {Patients}.
arXiv
(2018).
\doiurl{10.48550/arXiv.1804.03240} .
\url{http://arxiv.org/abs/1804.03240}
Accessed 2023-12-11
\end{botherref}
\endbibitem

\bibitem[\protect\citeauthoryear{Tahayori
  et~al.}{2021}]{tahayori_advanced_2021}
\begin{barticle}
\bauthor{\bsnm{Tahayori}, \binits{B.}},
\bauthor{\bsnm{Chini-Foroush}, \binits{N.}},
\bauthor{\bsnm{Akhlaghi}, \binits{H.}}:
\batitle{Advanced natural language processing technique to predict patient
  disposition based on emergency triage notes}.
\bjtitle{Emergency Medicine Australasia}
\bvolume{33}(\bissue{3}),
\bfpage{480}--\blpage{484}
(\byear{2021})
\doiurl{10.1111/1742-6723.13656} .
Accessed 2023-12-11
\end{barticle}
\endbibitem

\bibitem[\protect\citeauthoryear{Chen et~al.}{2022}]{chen2022imbalanced}
\begin{barticle}
\bauthor{\bsnm{Chen}, \binits{T.-L.}},
\bauthor{\bsnm{Chen}, \binits{J.C.}},
\bauthor{\bsnm{Chang}, \binits{W.-H.}},
\bauthor{\bsnm{Tsai}, \binits{W.}},
\bauthor{\bsnm{Shih}, \binits{M.-C.}},
\bauthor{\bsnm{Nabila}, \binits{A.W.}}:
\batitle{Imbalanced prediction of emergency department admission using natural
  language processing and deep neural network}.
\bjtitle{Journal of Biomedical Informatics}
\bvolume{133},
\bfpage{104171}
(\byear{2022})
\end{barticle}
\endbibitem

\bibitem[\protect\citeauthoryear{Wang et~al.}{2019}]{wang_deeptriager_2019}
\begin{bchapter}
\bauthor{\bsnm{Wang}, \binits{G.}},
\bauthor{\bsnm{Liu}, \binits{X.}},
\bauthor{\bsnm{Xie}, \binits{K.}},
\bauthor{\bsnm{Chen}, \binits{N.}},
\bauthor{\bsnm{Chen}, \binits{T.}}:
\bctitle{{DeepTriager}: {A} {Neural} {Attention} {Model} for {Emergency}
  {Triage} with {Electronic} {Health} {Records}}.
In: \bbtitle{2019 {IEEE} {International} {Conference} on {Bioinformatics} and
  {Biomedicine} ({BIBM})},
pp. \bfpage{978}--\blpage{982}
(\byear{2019}).
\doiurl{10.1109/BIBM47256.2019.8983093} .
\burl{https://ieeexplore.ieee.org/document/8983093}
Accessed 2023-12-11
\end{bchapter}
\endbibitem

\bibitem[\protect\citeauthoryear{Arik and Pfister}{2021}]{arik2021tabnet}
\begin{bchapter}
\bauthor{\bsnm{Arik}, \binits{S.{\"O}.}},
\bauthor{\bsnm{Pfister}, \binits{T.}}:
\bctitle{Tabnet: Attentive interpretable tabular learning}.
In: \bbtitle{Proceedings of the AAAI Conference on Artificial Intelligence},
vol. \bseriesno{35},
pp. \bfpage{6679}--\blpage{6687}
(\byear{2021})
\end{bchapter}
\endbibitem

\bibitem[\protect\citeauthoryear{Chen et~al.}{2022}]{chen_imbalanced_2022}
\begin{barticle}
\bauthor{\bsnm{Chen}, \binits{T.-L.}},
\bauthor{\bsnm{Chen}, \binits{J.C.}},
\bauthor{\bsnm{Chang}, \binits{W.-H.}},
\bauthor{\bsnm{Tsai}, \binits{W.}},
\bauthor{\bsnm{Shih}, \binits{M.-C.}},
\bauthor{\bsnm{Wildan~Nabila}, \binits{A.}}:
\batitle{Imbalanced prediction of emergency department admission using natural
  language processing and deep neural network}.
\bjtitle{Journal of Biomedical Informatics}
\bvolume{133},
\bfpage{104171}
(\byear{2022})
\doiurl{10.1016/j.jbi.2022.104171} .
Accessed 2023-12-11
\end{barticle}
\endbibitem

\end{thebibliography}


\end{document}